\PassOptionsToPackage{table}{xcolor}
\documentclass[10pt,twocolumn,letterpaper]{article}

\usepackage[pagenumbers]{iccv} 

\usepackage{amsmath,amsfonts,bm}

\def\Secref#1{Section~\ref{#1}}

\def\eqref#1{equation~\ref{#1}}

\def\1{\bm{1}}

\def\vmu{{\bm{\mu}}}

\def\vc{{\bm{c}}}

\def\ve{{\bm{e}}}

\def\vr{{\bm{r}}}
\def\vs{{\bm{s}}}
\def\vt{{\bm{t}}}

\def\mG{{\bm{G}}}

\def\mI{{\bm{I}}}

\def\mK{{\bm{K}}}

\def\mM{{\bm{M}}}

\def\mP{{\bm{P}}}

\def\mX{{\bm{X}}}
\def\mY{{\bm{Y}}}

\DeclareMathAlphabet{\mathsfit}{\encodingdefault}{\sfdefault}{m}{sl}
\SetMathAlphabet{\mathsfit}{bold}{\encodingdefault}{\sfdefault}{bx}{n}

\newcommand{\R}{\mathbb{R}}

\definecolor{iccvblue}{rgb}{0.21,0.49,0.74}
\usepackage[accsupp]{axessibility}
\usepackage[pagebackref,breaklinks,colorlinks,allcolors=iccvblue]{hyperref}
\usepackage{multirow}
\usepackage{cuted}
\usepackage{bbding}
\usepackage{capt-of}
\usepackage{nth}

\def\paperID{13546} 
\def\confName{ICCV}
\def\confYear{2025}

\title{FreeSplatter: Pose-free Gaussian Splatting for Sparse-view 3D Reconstruction}

\author{
Jiale Xu\textsuperscript{\rm 1} \quad
Shenghua Gao\textsuperscript{\rm 2} \quad
Ying Shan\textsuperscript{\rm 1}\\
\textsuperscript{\rm 1}ARC Lab, Tencent PCG \quad
\textsuperscript{\rm 2}The University of Hong Kong \\
{\tt\small \url{https://bluestyle97.github.io/projects/freesplatter/}}
}

\begin{document}
\maketitle

\begin{strip}
\vspace{-1.5cm}
\centering
\includegraphics[width=\textwidth]{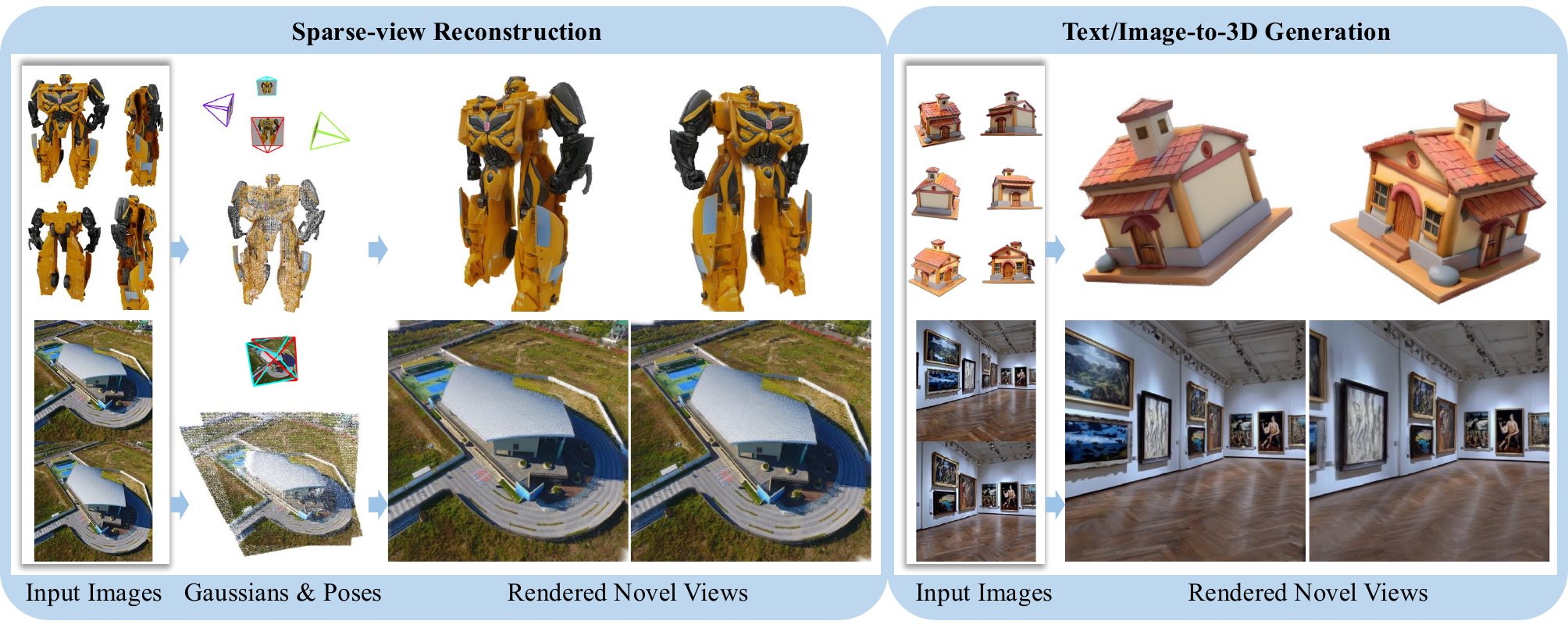}
\vspace{-6mm}
\captionof{figure}{FreeSplatter reconstructs high-fidelity 3D Gaussians and estimates accurate camera poses from uncalibrated sparse-view images in a feed-forward manner, handling both object-centric (\nth{1} row) and scene-level (\nth{2} row) scenarios effectively. It can also seamlessly process synthetic multi-view images from diffusion models, enabling efficient and high-quality text/image-to-3D content creation.}
\vspace{-2mm}
\label{fig:teaser}
\end{strip}

\begin{abstract}

Sparse-view reconstruction models typically require precise camera poses, yet obtaining these parameters from sparse-view images remains challenging. We introduce \textbf{FreeSplatter}, a scalable feed-forward framework that generates high-quality 3D Gaussians from \textbf{uncalibrated} sparse-view images while estimating camera parameters within seconds. Our approach employs a streamlined transformer architecture where self-attention blocks facilitate information exchange among multi-view image tokens, decoding them into pixel-aligned 3D Gaussian primitives within a unified reference frame. This representation enables both high-fidelity 3D modeling and efficient camera parameter estimation using off-the-shelf solvers. We develop two specialized variants--for \textbf{object-centric} and \textbf{scene-level} reconstruction--trained on comprehensive datasets. Remarkably, FreeSplatter outperforms several pose-dependent Large Reconstruction Models (LRMs) by a notable margin while achieving comparable or even better pose estimation accuracy compared to state-of-the-art pose-free reconstruction approach MASt3R in challenging benchmarks. Beyond technical benchmarks, FreeSplatter streamlines text/image-to-3D content creation pipelines, eliminating the complexity of camera pose management while delivering exceptional visual fidelity.
\end{abstract}    
\section{Introduction}
\label{sec:intro}

Recent breakthroughs in neural scene representation and differentiable rendering, \emph{e.g.}, Neural Radiance Fields (NeRF)\cite{mildenhall2021nerf} and Gaussian Splatting (GS)\cite{kerbl20233d}, have demonstrated exceptional multi-view reconstruction quality for densely-captured images with calibrated camera poses through per-scene optimization. However, these approaches fail in sparse-view scenarios where traditional camera calibration techniques like Structure-from-Motion (SfM)\cite{schonberger2016structure} struggle due to insufficient image overlaps. While generalizable reconstruction models\cite{hong2024lrm, xu2024instantmesh, charatan2024pixelsplat} address sparse-view reconstruction using learned priors in a feed-forward manner, they still require accurate camera parameters, sidestepping a fundamental challenge in real-world applications. Liberating sparse-view reconstruction from known camera poses remains a critical frontier.

Previous pose-free reconstruction efforts include PF-LRM~\cite{wang2024pflrm} and LEAP~\cite{jiang2024leap}, which map multi-view image tokens to NeRF representations using transformers. Despite their pioneering contributions, their approaches suffer from inefficient volume rendering and limited resolution, hampering training efficiency and scalability to complex scenes. Moreover, inferring camera poses from their implicit representations requires additional specialized components, introducing extra complexity. DUSt3R~\cite{wang2024DUSt3R} presents an alternative paradigm for joint 3D reconstruction and pose estimation through direct point regression, enabling efficient camera pose recovery with PnP solvers~\cite{fischler1981random, hartley2003multiple} and demonstrating impressive zero-shot capabilities.


However, point clouds' inherent sparsity limits their utility for downstream applications like novel view synthesis. In contrast, 3D Gaussian Splats (3DGS) can encode high-fidelity radiance fields while enabling efficient rendering by augmenting point clouds with additional attributes. This raises the question: can we directly predict ``Gaussian maps" from multi-view images to achieve both high-quality 3D modeling and instant camera pose estimation?

We introduce FreeSplatter, a feed-forward reconstruction framework that jointly predicts pixel-wise Gaussians from uncalibrated sparse-view images and estimates their camera parameters. At its core is a scalable streamlined transformer that maps multi-view image tokens into pixel-aligned Gaussian maps using simple self-attention layers—requiring no camera poses, intrinsics, or post-alignment. These Gaussian maps enable both high-fidelity scene representation and ultra-fast camera parameter estimation using off-the-shelf solvers~\cite{fischler1981random, hartley2003multiple, plastria2011weiszfeld}.


Leveraging the training and rendering efficiency of 3D Gaussians, we extend our approach to complex scene-level reconstruction by training two variants: FreeSplatter-O for object-centric reconstruction (trained on Objaverse~\cite{deitke2023objaverse}) and FreeSplatter-S for scene-level reconstruction (trained on mixed datasets~\cite{yao2020blendedmvs, yeshwanth2023scannet++, reizenstein2021common}). Both share a common architecture with task-specific adjustments. Our extensive experiments demonstrate FreeSplatter's superiority over existing methods in both reconstruction quality and pose estimation accuracy. Notably, FreeSplatter-O significantly outperforms several existing \emph{pose-dependent} large reconstruction models, while FreeSplatter-S achieves comparable or better pose estimation accuracy than state-of-the-art MASt3R~\cite{leroy2024grounding} across challenging benchmarks. We further demonstrate FreeSplatter's potential for enhancing 3D content creation pipelines through integration with multi-view diffusion models.

\section{Related Work}
\label{sec:rel}

\noindent\textbf{Large Reconstruction Models.}
Large-scale 3D object datasets~\cite{deitke2023objaverse, deitke2024objaverse} have enabled training of highly generalizable models for open-category image-to-3D reconstruction. Large Reconstruction Models (LRMs)~\cite{hong2024lrm, xu2024dmvd, li2024instantd} employ scalable feed-forward transformer architectures to map sparse-view image tokens into 3D triplane NeRF representations~\cite{mildenhall2021nerf, chan2022efficient}, supervised with multi-view rendering losses. Recent advances have explored alternative representations including meshes~\cite{xu2024instantmesh, wei2024meshlrm, wang2024crm} and 3D Gaussians~\cite{tang2024lgm, xu2024grm, zhang2024gs} for real-time rendering, more efficient network architectures~\cite{zheng2024mvd, zhang2024geolrm, chen2024lara, li2024m, cao2024lightplane}, enhanced texture quality~\cite{boss2024sf3d, siddiqui2024meta, yang2024magic}, and explicit 3D supervision for improved geometry~\cite{liu2024meshformer}. Despite their impressive reconstruction quality and generalization capabilities, LRMs require \emph{posed} images as input and are highly sensitive to pose accuracy, significantly limiting their practical application scenarios.

\begin{figure*}[t]
\centering
\includegraphics[width=0.9\textwidth]{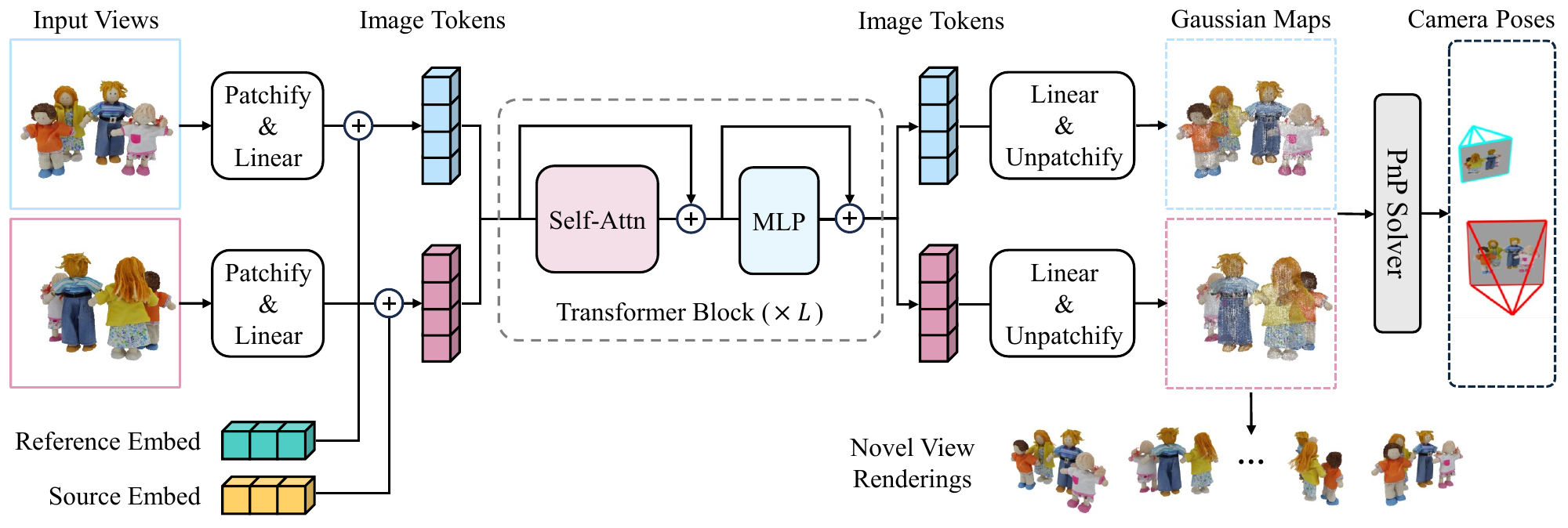}
\vspace{-3mm}
\caption{\textbf{FreeSplatter Pipeline.} Given $N$ uncalibrated input views without any known camera extrinsics or intrinsics, we first patchify each image into tokens and feed these tokens into a sequence of self-attention blocks, enabling information exchange across multiple views. The resulting tokens are then decoded into $N$ Gaussian maps, which allow us to render novel views and simultaneously recover the camera focal length $f$ and poses using simple iterative solvers.}
\label{fig:pipeline}
\vspace{-5mm}
\end{figure*}

\noindent\textbf{Pose-free Reconstruction.} 
Classical pose-free reconstruction algorithms like Structure from Motion (SfM)~\cite{hartley2003multiple, ullman1979interpretation, schonberger2016structure} first establish pixel correspondences across multiple views, then perform 3D point triangulation and bundle adjustment to jointly optimize 3D coordinates and camera parameters. Recent improvements to SfM leverage learning-based feature descriptors~\cite{detone2018superpoint, dusmanu2019d2, revaud2019r2d2, wang2023guiding}, image matchers~\cite{edstedt2023dkm, edstedt2024roma, sarlin2020superglue, lindenberger2023lightglue}, and differentiable bundle adjustment~\cite{lin2021barf, wang2023visual, wei2020deepsfm}. While effective with sufficient image overlaps, SfM-based methods struggle with sparse views where correspondence matching becomes challenging. Learning-based methods~\cite{jiang2024few, jiang2024leap, hong2024unifying, fan2023pose} address this by utilizing learned data priors to recover 3D geometry from input views. PF-LRM~\cite{wang2024pflrm} extends the LRM framework by predicting per-view coarse point clouds for camera pose estimation with a differentiable PnP solver~\cite{chen2022epro}. DUSt3R~\cite{wang2024DUSt3R} introduces a novel approach to Multi-view Stereo (MVS) by framing it as a pointmap-regression problem, with subsequent works enhancing its local representation capabilities~\cite{leroy2024grounding} and reconstruction efficiency~\cite{wang20243d}.

\noindent\textbf{Generalizable Gaussian Splatting.}
Compared to the implicit MLP-based representation of NeRF~\cite{mildenhall2021nerf}, 3D Gaussian Splatting (3DGS)~\cite{kerbl20233d, huang20242d} explicitly represents scenes as point clouds with additional attributes, achieving an balance between high-fidelity rendering and real-time performance. However, traditional 3DGS requires per-scene optimization with densely-captured images and SfM-generated sparse point clouds for initialization. Recent research~\cite{charatan2024pixelsplat, chen2024mvsplat, szymanowicz2024splatter, liu2024fast, wewer2024latentsplat} has explored feed-forward models for sparse-view Gaussian reconstruction by leveraging large-scale datasets and scalable architectures. These approaches typically assume access to accurate camera poses and employ 3D-to-2D geometric projection for feature aggregation, using techniques like epipolar lines~\cite{charatan2024pixelsplat} or plane-swept cost volumes~\cite{chen2024mvsplat, liu2024fast}. InstantSplat~\cite{fan2024instantsplat} and Splatt3R~\cite{smart2024splatt3r} leverage DUSt3R/MASt3R's pose-free reconstruction capabilities—the former initializes Gaussian positions using DUSt3R point clouds before optimizing other Gaussian parameters, while the latter trains a Gaussian head on a frozen MASt3R model. Despite impressive results, their reconstruction quality remains heavily dependent on the quality of the initial point clouds generated by DUSt3R.

\section{Method}
\label{sec:method}

Given $N$ input images $\left\{\mI^n \mid n=1,\ldots,N\right\}$ without known camera parameters, FreeSplatter performs joint scene reconstruction and camera parameter estimation. The pipeline is formulated as:
\begin{equation}
\mG, \mP^1, \ldots, \mP^N, f = \operatorname{FreeSplatter}\left(\mI^1, \ldots, \mI^N\right),
\label{eq:pipeline}
\end{equation}
where $\mG = \left\{\mG^n \mid n=1,\ldots,N\right\}$ represents the unified set of reconstructed Gaussian maps, $\mP^n$ denotes the estimated camera pose for $\mI^n$, and $f$ represents the shared focal length across views (reasonable in most scenarios).


\subsection{Preliminary}
\label{sec:method:pre}

3D Gaussian Splatting (3DGS)~\cite{kerbl20233d} represents a scene as a set of 3D Gaussian primitives. Each primitive is parameterized by location $\vmu_k \in \R^3$, rotation quaternion $\vr_k \in \R^4$, scale $\vs_k \in \R^3$, opacity $o_k \in \R$, and Spherical Harmonic (SH) coefficients $\vc_k \in \R^{3\times d^2}$ for computing view-dependent color ($d$ denoting the degree of SH). This representation parameterizes scene radiance fields through explicit point clouds, enabling efficient novel view synthesis via rasterization. Compared to NeRF's computationally intensive volume rendering, 3DGS achieves comparable visual quality with significantly reduced computational and memory requirements.


\subsection{Model Architecture}
\label{sec:method:model}

As Figure~\ref{fig:pipeline} shows, FreeSplatter adopts a transformer architecture inspired by GS-LRM~\cite{zhang2024gs}. For input images $\left\{\mI^n \mid n=1,\ldots,N\right\}$, the model patchifies them into tokens $\left\{\ve^{n,m} \mid n=1,\ldots,N,m=1,\ldots,M\right\}$ ($M$ denotes patch number per image), processes them through self-attention blocks for multi-view information exchange, and decodes them into $N$ Gaussian maps $\left\{\mG^n \mid n=1,\ldots,N\right\}$. These maps enable novel view synthesis and camera parameter recovery through iterative optimization.

\noindent\textbf{Image Tokenization.}
The model processes $N$ input images $\left\{\mI^n \in \R^{H \times W \times 3} \mid n=1, \ldots, N\right\}$ using ViT-style~\cite{dosovitskiy2021an} tokenization: images are divided into $p\times p$ patches ($p=8$), flattened to $p^2 \cdot 3$ dimensional vectors, and projected to $d$-dimensional tokens via a linear layer. Each token $\ve^{n,m}$ is enhanced with position and view embeddings:
\begin{equation}
    \ve^{n,m} = \ve^{n,m} + \ve^m_{\mathrm{pos}} + \ve^n_{\mathrm{view}},
    \label{eq:embed}
\end{equation}
where $\ve^m_{\mathrm{pos}}$ encodes patch position and $\ve^n_{\mathrm{view}}$ distinguishes reference and source views. Specifically, we take the first image as the reference view and predict all Gaussian in its camera frame. We use a reference embedding $\ve^{\mathrm{ref}}$ for the first view ($n=1$) and another source embedding $\ve^{\mathrm{src}}$ for other views ($n=2, \ldots, N$), both of which are learnable.

\noindent\textbf{Feed-forward Transformer.}
The augmented tokens undergo processing through $L$ self-attention blocks, each combining self-attention and MLP layers with pre-normalization and residual connections~\cite{he2016deep}. 

\begin{figure*}[t]
\centering
\includegraphics[width=1.0\textwidth]{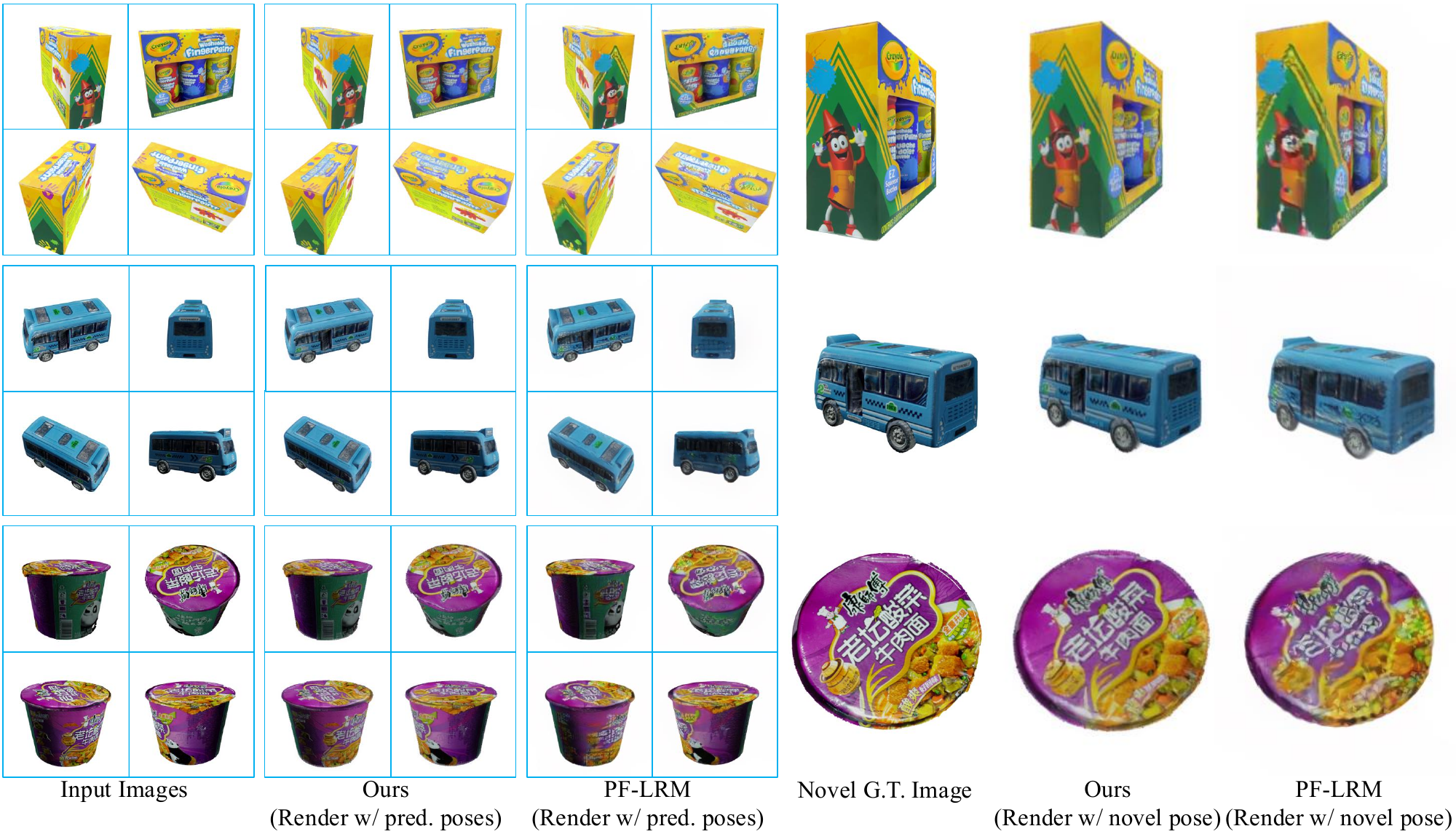}
\vspace{-6mm}
\caption{\textbf{Sparse-view Reconstruction on PF-LRM's Evaluation Datasets.} FreeSplatter-O synthesizes significantly better visual details than PF-LRM. The \nth{1} row is from the GSO dataset, while the \nth{2} and \nth{3} rows are from the OmniObject3D dataset.}
\vspace{-5mm}
\label{fig:comparison_pflrm_1}
\end{figure*}

\noindent\textbf{Gaussian Map Prediction.} 
Each image token $\ve^{n,m}_{\mathrm{out}}$ outputed by the last self-attention block is transformed into $p^2$ Gaussians with a linear layer, yielding vectors of dimension $p^2\cdot q$ ($q$ being the Gaussian parameter count). These predictions are reshaped into Gaussian patches $\mG^{n,m}\in \R^{p\times p\times q}$ and spatially concatenated to form $N$ Gaussian maps $\left\{\mG^n \in \R^{H\times W\times q} \mid n=1,\ldots,N\right\}$.

Each map pixel represents a $q$-dimensional 3D Gaussian primitive. Unlike pose-dependent Gaussian LRMs~\cite{zhang2024gs, xu2024grm, tang2024lgm} that use single depth values for Gaussian locations, our pose-free setting precludes depth-based unprojection. Instead, we directly predict Gaussian locations in the reference frame and enforce pixel alignment through a dedicated loss term to restrict Gaussians to lie on camera rays (detailed in \Secref{sec:method:train}).

\noindent\textbf{Camera Pose Estimation.}
Camera parameter recovery begins with focal length $f$ estimation from predicted Gaussian maps. Unlike DUSt3R, which requires \emph{pairwise} image processing and subsequent \emph{global alignment}, FreeSplatter predicts all Gaussian maps in a unified reference frame, enabling direct camera pose estimation for all views. Given the $n$-th view's Gaussian location map $\mX^n \in \R^{H\times W\times 3}$ (first 3 channels of $\mG^n$), corresponding pixel coordinate map $\mY^n \in \R^{H\times W\times 2}$, and validity mask $\mM^n \in \R^{H\times W}$, we employ PnP-RANSAC~\cite{hartley2003multiple, opencv_library} to compute the camera pose $\mP^n \in \R^{4\times 4}$:
\begin{equation}
    \mP^n=\operatorname{PnP}\left(\mX^n, \mY^n, \mM^n, \mK\right),
    \label{eq:pnp}
\end{equation}
where $\mK=[[f,0,\frac{W}{2}], [0,f,\frac{H}{2}], [0,0,1]]$ represents the estimated intrinsic matrix. The mask $\mM^n$ identifies valid pixels for pose optimization, which is implemented differently for object-centric and scene-level reconstruction. Please refer to Section 1 of the supplementary material for more implementation details.


\subsection{Training Details}
\label{sec:method:train}

FreeSplatter offers two variants optimized for object-centric and scene-level pose-free reconstruction. While sharing architectural elements and parameter scale, these variants employ distinct training objectives and strategies.

\noindent\textbf{Two-stage Training Strategy.}
Prior pose-dependent LRMs leverage pure rendering loss for supervision~\cite{hong2024lrm, li2024instantd, xu2024grm, zhang2024gs}. However, our model assumes no known camera poses nor intrinsics and the Gaussian positions are free in 3D space, making it extremely challenging to predict correct Gaussian positions. Gaussian-based reconstruction approaches heavily rely on the initialization of Gaussian positions, \emph{e.g.}, 3DGS~\cite{kerbl20233d} initializes the Gaussian positions with the sparse point cloud generated by SfM, while the parameters of our model are randomly initialized at the beginning. In practice, we found it essential to supervise the Gaussian positions at the beginning:
\begin{equation}
    \mathcal{L}_{\mathrm{pos}} = \sum_{n=1}^N \left\|\mM^n \odot \hat{\mX}^n - \mM^n \odot \mX^n\right\|,
    \label{eq:loss:position}
\end{equation}
where $\hat{\mX}^n\in \R^{H\times W\times 3}$ represents predicted positions, $\mX^n$ denotes ground truth positions from depth unprojection, and $\mM^n \in \R^{H\times W}$ masks valid depth values, which is the foreground object mask for object-centric reconstruction. For scene-level reconstruction, $\mM^n$ depends on where the depth values are defined in different datasets.

We apply $\mathcal{L}_{\mathrm{pos}}$ in the pre-training stage, so that the model learns to predict approximately correct Gaussian positions. In our experiments, this pre-training is \emph{essential} to model's convergence. However, $\mathcal{L}_{\mathrm{pos}}$ can only supervise the pixels with valid depths, while the Gaussian positions predicted at other pixels remain unconstrained. Besides, the ground truth depths are noisy in some datasets, and applying $\mathcal{L}_{\mathrm{pos}}$ throughout the training leads to degraded rendering quality. To provide a more stable geometric supervision, we adopt a pixel-alignment loss to enforce each predicted Gaussian to be aligned with its corresponding pixel through cosine similarity maximization:
\begin{equation}
    \mathcal{L}_{\mathrm{align}} = \sum_{n=1}^N \sum_{i=0}^{H-1} \sum_{j=0}^{W-1} \left(1- \frac{\hat{\vr}^n_{i,j} \cdot \vr^n_{i,j}}{\|\hat{\vr}^n_{i,j}\| \|\vr^n_{i,j}\|} \right),
    \label{eq:loss:align}
\end{equation}
where $\vr^n_{i,j}$ denotes the ray from the camera origin $\vt^n$ to point $\mX^n_{i,j}$. $\mathcal{L}_{\mathrm{align}}$ restricts the predicted Gaussians to be distributed on the camera rays, which enhances rendering quality and facilitates camera parameter estimation by minimizing pixel-projection errors.

\noindent\textbf{Loss Functions.}
The overall training objective is:
\begin{equation}
    \mathcal{L} = \mathcal{L}_{\mathrm{render}} + \lambda_\mathrm{a} \cdot \mathcal{L}_\mathrm{align} + \1_\mathrm{t\le T_\mathrm{max}} \lambda_\mathrm{p} \cdot \mathcal{L}_\mathrm{pos},
    \label{eq:loss:all}
\end{equation}
where the rendering loss $\mathcal{L}_{\mathrm{render}}$ is a combination of MSE and LPIPS loss. $t$ and $T_\mathrm{max}$ denote the current training step and maximum pre-training step, respectively. In our implementation, we set $\lambda_\mathrm{a}=1.0, \lambda_\mathrm{p}=10.0, T_\mathrm{max}=10^5$.

\begin{figure}[t]
\centering
\includegraphics[width=1.0\columnwidth]{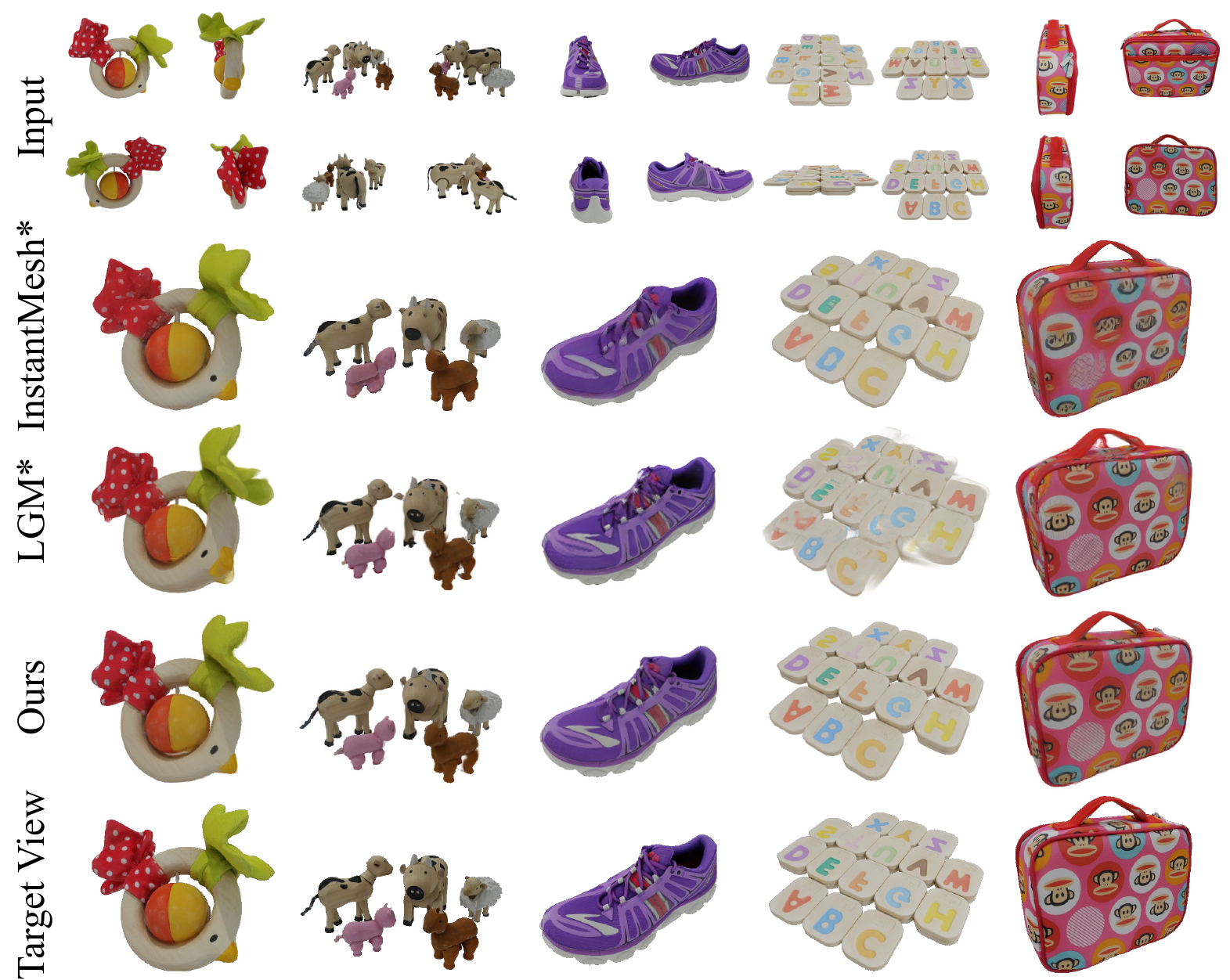}
\vspace{-6mm}
\caption{\textbf{Sparse-view Reconstruction on GSO dataset.} * indicates that ground truth camera poses are used as input.}
\label{fig:comparison_obj}
\vspace{-5mm}
\end{figure}

\noindent\textbf{Occlusion in Pixel-aligned Gaussians.}
Pose-dependent Gaussian-based LRMs~\cite{tang2024lgm, zhang2024gs, xu2024grm} parameterize Gaussian positions with single depth values to ensure pixel alignment. Despite the simplicity, this approach limits reconstruction to areas directly observed in input views, potentially missing occluded regions in sparse-view scenarios. Our model addresses this limitation differently for object-centric and scene-level reconstruction: (i) For object-centric reconstruction, we apply $\mathcal{L}_\mathrm{align}$ (Equation~\ref{eq:loss:align}) exclusively to foreground regions, allowing Gaussians outside these areas to position freely and model occluded regions. (ii) For scene-level reconstruction with real-world imagery, complete pixel alignment is necessary to handle complex backgrounds. We focus on reconstructing observed areas and adopt Splatt3R's~\cite{smart2024splatt3r} target-view masking strategy, computing rendering loss only for visible regions to prevent negative training guidance from occluded areas.

\section{Experiments}
\label{sec:exp}

\begin{table}[t]
\scriptsize
\renewcommand{\tabcolsep}{1.0mm}
\renewcommand{\arraystretch}{1.0}
\centering
\resizebox{1.0\columnwidth}{!}{
\begin{tabular}{lcccccc}
\toprule
\multirow{2}{*}{\textbf{Method}} & 
\multicolumn{3}{c}{\textbf{GSO}} & 
\multicolumn{3}{c}{\textbf{OmniObject3D}} \\
\cmidrule(r){2-4} \cmidrule(r){5-7}
& 
PSNR $\uparrow$ & SSIM $\uparrow$ & LPIPS $\downarrow$ & 
PSNR $\uparrow$ & SSIM $\uparrow$ & LPIPS $\downarrow$ \\
\midrule 
& 
\multicolumn{6}{c}{Evaluate renderings at G.T. novel-view poses} \\
\midrule
PF-LRM & \cellcolor{cyan!20}25.08 & \cellcolor{cyan!20}0.877 & \cellcolor{cyan!20}0.095 & 21.77 & 0.866 & 0.097 \\
\textbf{FreeSplatter-O} & 23.54 & 0.864 & 0.100 & \cellcolor{cyan!20}22.83 & \cellcolor{cyan!20}0.876 & \cellcolor{cyan!20}0.088 \\
\midrule 
& 
\multicolumn{6}{c}{Evaluate renderings at predicted input poses} \\
\midrule 
PF-LRM & \cellcolor{cyan!20}27.10 & \cellcolor{cyan!20}0.905 & \cellcolor{cyan!20}0.065 & 25.86 & 0.901 & 0.062 \\
\textbf{FreeSplatter-O} & 25.50 & 0.897 & 0.076 & \cellcolor{cyan!20}26.49 & \cellcolor{cyan!20}0.926 & \cellcolor{cyan!20}0.050 \\
\bottomrule
\end{tabular}
}
\vspace{-3mm}
\caption{\textbf{Sparse-view Reconstruction on PF-LRM's Eval Data.}}
\vspace{-3mm}
\label{tab:pflrm_recon}
\end{table}

\begin{table}[t]
\scriptsize
\renewcommand{\tabcolsep}{2.2mm}
\renewcommand{\arraystretch}{1.0}
\centering
\resizebox{1.0\columnwidth}{!}{
\begin{tabular}{lcccc}
\toprule
\multirow{2}{*}{\textbf{Method}} & 
\multicolumn{4}{c}{\textbf{GSO}} \\
\cmidrule(r){2-5}
& RRE $\downarrow$ & RRA@$15^{\circ} \uparrow$ & RRA@$30^{\circ} \uparrow$ & TE$\downarrow$ \\
\midrule 
PF-LRM & \cellcolor{cyan!20}3.99 & \cellcolor{cyan!20}0.956 & \cellcolor{cyan!20}0.976 & \cellcolor{cyan!20}0.041 \\
\textbf{FreeSplatter-O} & 8.96 & 0.909 & 0.936 & 0.090 \\

\midrule 
& \multicolumn{4}{c}{\textbf{OmniObject3D}} \\
\cmidrule(r){2-5}
& 
RRE $\downarrow$ & RRA@$15^{\circ} \uparrow$ & RRA@$30^{\circ} \uparrow$ & TE$\downarrow$ \\
\midrule 
PF-LRM & 8.013 & 0.889 & 0.954 & 0.089 \\
\textbf{FreeSplatter-O} & \cellcolor{cyan!20}3.446 & \cellcolor{cyan!20}0.982 & \cellcolor{cyan!20}0.996 & \cellcolor{cyan!20}0.039 \\

\bottomrule
\end{tabular}
}
\vspace{-3mm}
\caption{\textbf{Camera Pose Estimation on PF-LRM's Eval Data.}}
\vspace{-6mm}
\label{tab:pflrm_pose}
\end{table}

We evaluate our method on both sparse-view reconstruction (Section \ref{sec:exp:recon}) and camera pose estimation (Section \ref{sec:exp:pose}) tasks, including object-centric and scene-level scenarios. Please refer to the supplementary material for additional implementation details and experimental results.

\subsection{Experimental Settings}
\label{sec:exp:setting}


\begin{figure*}[t]
\centering
\includegraphics[width=0.918\textwidth]{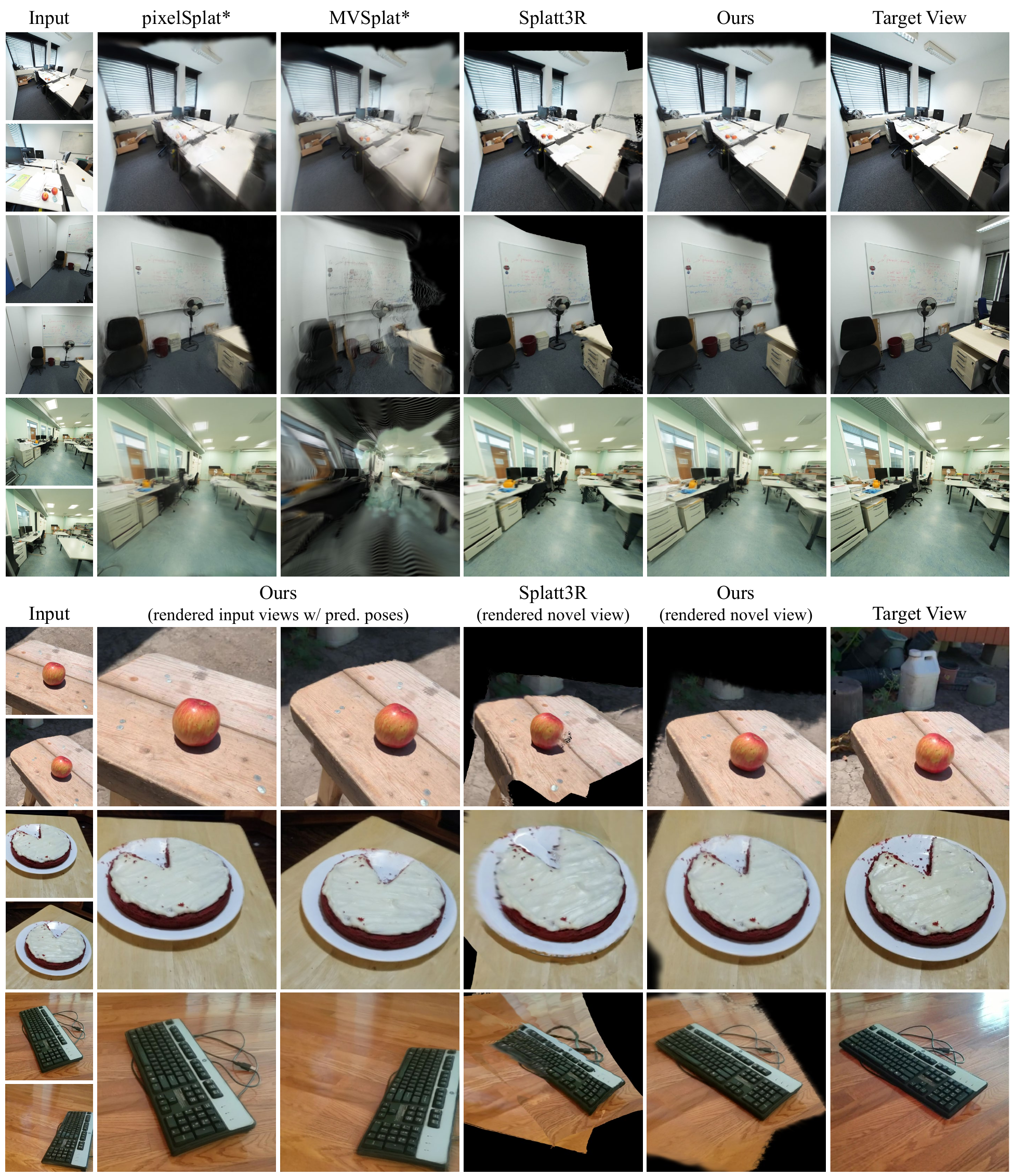}
\vspace{-3mm}
\caption{\textbf{Sparse-view Reconstruction on ScanNet++ (top) and CO3Dv2 (bottom).} * indicates that ground truth camera poses are used as input.}
\label{fig:comparison_scene}
\vspace{-5mm}
\end{figure*}

\noindent\textbf{Training Datasets.}
FreeSplatter-O is trained on Objaverse~\cite{deitke2023objaverse}, utilizing white-background renders of centered objects. Each 3D asset is normalized to a $[-1, 1]^3$ cube, with 32 randomly sampled views (with diverse camera intrinsics, \emph{i.e.}, focal lengths) and corresponding depth maps rendered at 512×512 resolution. FreeSplatter-S leverages a diverse training set comprising BlendedMVS~\cite{yao2020blendedmvs}, ScanNet++\cite{yeshwanth2023scannet++}, and CO3Dv2\cite{reizenstein2021common}—a subset of DUSt3R's~\cite{wang2024DUSt3R} training data encompassing outdoor scenes, indoor environments, and real-world objects.

\noindent\textbf{Evaluation Datasets.}
For object-level experiments, we utilize Google Scanned Objects (GSO)\cite{downs2022google} and OmniObject3D~\cite{wu2023omniobject3d} (chosen 300 objects across 30 categories). Each object is captured through 24 views: 20 random and 4 structured input views, the latter positioned uniformly at $20^\circ$ elevation for comprehensive coverage. In addition, we also use the GSO/OmniObject3D evaluation data provided by PF-LRM for comparison, since we can only access its inference results. Scene-level performance is assessed on the test splits of ScanNet++\cite{yeshwanth2023scannet++} and CO3Dv2~\cite{reizenstein2021common}.

\subsection{Sparse-view Reconstruction}
\label{sec:exp:recon}

\noindent\textbf{Baselines.} 
Prior pose-free object reconstruction approaches like LEAP~\cite{jiang2024leap} exhibits limited generalization due to its small-scale training, while PF-LRM~\cite{wang2024pflrm} is highly relevant and serves as our baseline for both object-level reconstruction and pose estimation tasks. We also evaluate against two pose-dependent methods LGM~\cite{tang2024lgm} and InstantMesh~\cite{xu2024instantmesh}, which leverage 3D Gaussians and tri-plane NeRF respectively, using ground truth camera poses.
For scene-level reconstruction, we compare against two state-of-the-art generalizable Gaussian methods: pixelSplat~\cite{charatan2024pixelsplat} and MVSplat~\cite{chen2024mvsplat}. Both methods are fine-tuned on ScanNet++ after pre-training on RealEstate10K~\cite{zhou2018stereo}. We also evaluate against Splatt3R~\cite{smart2024splatt3r}, a pose-free approach that combines a frozen MASt3R~\cite{leroy2024grounding} backbone with a trainable head for Gaussian attribute prediction.


\noindent\textbf{Metrics.} 
We evaluate the performance of sparse-view reconstruction using standard novel view synthesis metrics (PSNR, SSIM, and LPIPS) at 512×512 resolution.

\begin{table}[t]
\scriptsize
\renewcommand{\tabcolsep}{1.0mm}
\renewcommand{\arraystretch}{1.0}
\centering
\resizebox{1.0\columnwidth}{!}{
\begin{tabular}{lcccccc}
\toprule
\multirow{2}{*}{\textbf{Method}} & 
\multicolumn{3}{c}{\textbf{GSO}} &
\multicolumn{3}{c}{\textbf{OmniObject3D}} \\
\cmidrule(r){2-4} \cmidrule(r){5-7}
& PSNR $\uparrow$ & SSIM $\uparrow$ & LPIPS $\downarrow$ & PSNR $\uparrow$ & SSIM $\uparrow$ & LPIPS $\downarrow$ \\
\midrule 
LGM* & 24.463 & 0.891 & 0.093 & 24.852 & 0.942 & 0.060 \\
InstantMesh* & 25.421 & 0.891 & 0.095 & 24.077 & 0.945 & 0.062 \\
\textbf{FreeSplatter-O} & \cellcolor{cyan!20}30.443 & \cellcolor{cyan!20}0.945 & \cellcolor{cyan!20}0.055 & \cellcolor{cyan!20}31.929 & \cellcolor{cyan!20}0.973 & \cellcolor{cyan!20}0.027 \\

\midrule 
\multirow{2}{*}{\textbf{Method}} & 
\multicolumn{3}{c}{\textbf{ScanNet++}} & 
\multicolumn{3}{c}{\textbf{CO3Dv2}} \\ 
\cmidrule(r){2-4} \cmidrule(r){5-7}
& PSNR $\uparrow$ & SSIM $\uparrow$ & LPIPS $\downarrow$ & PSNR $\uparrow$ & SSIM $\uparrow$ & LPIPS $\downarrow$ \\
\midrule 
pixelSplat* & 24.974 & \cellcolor{cyan!20}0.889 & 0.180 & - & - & - \\
MVSplat* &  22.601 & 0.862 & 0.208 & - & - & - \\
Splatt3R & 21.013 & 0.830 & 0.209 & 18.074 & 0.740 & 0.197 \\
\textbf{FreeSplatter-S} & \cellcolor{cyan!20}25.807 & 0.887 & \cellcolor{cyan!20}0.140 & \cellcolor{cyan!20}20.405 & \cellcolor{cyan!20}0.781 & \cellcolor{cyan!20}0.162 \\

\bottomrule
\end{tabular}
}
\vspace{-3mm}
\caption{\textbf{Sparse-view Reconstruction on Object-centric and Scene-level Datasets.} We did not test pixelSplat/MVSplat on CO3Dv2 due to the significant domain gap. * indicates that ground truth camera poses are used as input.}
\label{tab:recon}
\vspace{-4mm}
\end{table}

\noindent\textbf{Comparison with PF-LRM.} 
Due to the lack of code, we benchmark against PF-LRM using their provided evaluation datasets and inference results. As Table~\ref{tab:pflrm_recon} shows, while PF-LRM achieves superior metrics on their GSO evaluation dataset, FreeSplatter-O performs better on their OmniObject3D evaluation dataset. This disparity can be attributed to PF-LRM's GSO evaluation images being rendered under identical conditions (\emph{e.g.}, light intensity, camera distribution) as their training data, whereas OmniObject3D uses original dataset images, providing a more objective comparison. Qualitative results in Figure~\ref{fig:comparison_pflrm_1} demonstrate FreeSplatter's superior preservation of visual details.

\noindent\textbf{Comparison with Pose-dependent LRMs.} 
On our object-centric evaluation datasets, FreeSplatter-O significantly outperforms pose-dependent methods LGM and InstantMesh, achieving PSNR improvements of $>5$ and $>7$ on GSO and OmniObject3D respectively, despite their usage of ground truth camera poses (Table~\ref{tab:recon}). Qualitative comparisons in Figure~\ref{fig:comparison_obj} reveal superior detail preservation by our method, particularly evident in text rendering (4th column), while competitors exhibit blurring artifacts. Existing works on LRMs assume the necessity of accurate camera poses for high-quality 3D reconstruction, incorporating pose information through LayerNorm modulation~\cite{li2024instantd} or plucker ray embeddings~\cite{xu2024dmvd, tang2024lgm, xu2024grm}. However, FreeSplatter-O's superior performance suggests that scalable and high-quality sparse-view reconstruction is feasible without known accurate camera poses in certain cases.

\noindent\textbf{Results on Scene-level Reconstruction.} 
For scene-level reconstruction, FreeSplatter-S outperforms pose-dependent methods (pixelSplat, MVSplat) on most ScanNet++ metrics (Table~\ref{tab:recon}). While Splatt3R, a pose-free alternative, employs MASt3R's~\cite{leroy2024grounding} frozen architecture for point prediction, its performance is limited by fixed Gaussian positions. Our end-to-end training approach enables joint optimization of Gaussian parameters, resulting in superior visual fidelity on both ScanNet++ and CO3Dv2 datasets (Figure~\ref{fig:comparison_scene}). To be noted, novel view synthesis for pose-free methods is accomplished through camera alignment with target viewpoints.


\begin{table}[t]
\scriptsize
\renewcommand{\tabcolsep}{2.2mm}
\renewcommand{\arraystretch}{1.0}
\centering
\resizebox{1.0\columnwidth}{!}{
\begin{tabular}{lcccc}
\toprule

\multirow{2}{*}{\textbf{Method (Object)}} & \multicolumn{4}{c}{\textbf{GSO}} \\
\cmidrule(r){2-5}
& RRE $\downarrow$ & RRA@$15^{\circ} \uparrow$ & RRA@$30^{\circ}\uparrow$ & TE$\downarrow$ \\
\midrule 
FORGE  & 97.814 & 0.022 & 0.083 & 0.898 \\
MASt3R & 59.633 & 0.269 & 0.456 & 0.353 \\
\textbf{FreeSplatter-O} & \cellcolor{cyan!20}3.902 & \cellcolor{cyan!20}0.961 & \cellcolor{cyan!20}0.977 & \cellcolor{cyan!20}0.040 \\

\midrule 

& \multicolumn{4}{c}{\textbf{OmniObject3D}} \\
\cmidrule(r){2-5}
& RRE $\downarrow$ & RRA@$15^{\circ} \uparrow$ & RRA@$30^{\circ}\uparrow$ & TE$\downarrow$ \\
\midrule
FORGE & 76.822 & 0.081 & 0.257 & 0.430 \\
MASt3R & 91.204 & 0.105 & 0.212 & 0.524 \\
\textbf{FreeSplatter-O} & \cellcolor{cyan!20}11.346 & \cellcolor{cyan!20}0.909 & \cellcolor{cyan!20}0.935 & \cellcolor{cyan!20}0.104 \\

\midrule 

\multirow{2}{*}{\textbf{Method (Scene)}} & \multicolumn{4}{c}{\textbf{ScanNet++}} \\
\cmidrule(r){2-5}
& RRE $\downarrow$ & RRA@$15^{\circ} \uparrow$ & RRA@$30^{\circ}\uparrow$ & TE$\downarrow$ \\
\midrule 

RoMa & 0.862 & 0.977 & 0.985 & 0.421 \\
MASt3R & \cellcolor{cyan!20}0.724 & 0.988 & \cellcolor{cyan!20}0.993 & 0.356 \\
\textbf{FreeSplatter-S} & 0.776 & \cellcolor{cyan!20}0.991 & 0.990 & \cellcolor{cyan!20}0.066 \\

\midrule 
 & \multicolumn{4}{c}{\textbf{CO3Dv2}} \\
\cmidrule(r){2-5}
& RRE $\downarrow$ & RRA@$15^{\circ} \uparrow$ & RRA@$30^{\circ}\uparrow$ & TE$\downarrow$ \\
\midrule 
PoseDiffusion & 7.950 & 0.803 & 0.868 & 0.409 \\
RayDiffusion & 7.028 & 0.833 & 0.890 & 0.482 \\
RoMa & 5.377 & 0.839 & 0.922 & 0.335 \\
MASt3R & \cellcolor{cyan!20}2.917 & 0.975 & \cellcolor{cyan!20}0.989 & 0.299 \\
\textbf{FreeSplatter-S} & 3.048 & \cellcolor{cyan!20}0.976 & 0.986 & \cellcolor{cyan!20}0.190 \\

\midrule 
 & \multicolumn{4}{c}{\textbf{Re10K}} \\
\cmidrule(r){2-5}
& RRE $\downarrow$ & RRA@$15^{\circ} \uparrow$ & RRA@$30^{\circ}\uparrow$ & TE$\downarrow$ \\
\midrule 
PoseDiffusion & 14.387 & 0.732 & 0.780 & 0.466 \\
RayDiffusion & 12.023 & 0.767 & 0.814 & 0.439 \\
RoMa & 5.663 & 0.918 & 0.947 & 0.402 \\
MASt3R & \cellcolor{cyan!20}2.341 & 0.972 & 0.994 & 0.374 \\
\textbf{FreeSplatter-S} & 3.513 & \cellcolor{cyan!20}0.982 & \cellcolor{cyan!20}0.995 & \cellcolor{cyan!20}0.293 \\

\bottomrule
\end{tabular}
}
\vspace{-3mm}
\caption{\textbf{Camera Pose Estimation on Object-centric and Scene-level Datasets.} To be noted, Re10K is outside the training dataset.}
\label{tab:pose}
\vspace{-6mm}
\end{table}

\subsection{Camera Pose Estimation}
\label{sec:exp:pose}

\noindent\textbf{Baselines.}
We fisrt evaluate pose estimation performance against PF-LRM on its evaluation datasets. For all of our evaluation datasets, we benchmark against MASt3R, the current state-of-the-art in zero-shot multi-view pose estimation.  Additional comparisons include FORGE~\cite{jiang2024few} for object-centric evaluation, and PoseDiffusion~\cite{wang2023posediffusion}, RayDiffusion~\cite{zhang2024cameras}, and RoMa~\cite{edstedt2024roma} for scene-level tasks, with the former two excluded from ScanNet++ evaluation due to training scope limitations. Traditional COLMAP-based methods~\cite{schonberger2016structure} are omitted due to documented high failure rates in sparse-view scenarios~\cite{wang2024pflrm}. We further incorporate RealEstate10K~\cite{zhou2018stereo} (Re10K) test splits to assess generalization to challenging scenes.

\noindent\textbf{Metrics.}
Following established protocols~\cite{wang2024pflrm,wang2023posediffusion}, we evaluate pose estimation performance using both rotation and translation metrics: relative rotation error (RRE) in degrees, relative rotation accuracy (RRA) at $15^\circ$ and $30^\circ$ thresholds, and translation error (TE) measured as the distance between predicted and ground truth camera centers. For multi-view settings, errors are averaged over all possible pairs of cameras. It is important to note that the TE metric is scale-invariant: we first compute the relative translations between views for both ground truth and predictions, \emph{normalize} these translations by their respective mean $\ell_2$-norm, and then report the mean difference.

\noindent\textbf{Comparison with PF-LRM.}
Pose estimation results mirror reconstruction performance trends: PF-LRM excels on their GSO evaluation set, while FreeSplatter-O demonstrates superior performance on OmniObject3D. As we have analyzed, this disparity likely stems from PF-LRM's GSO evaluation images sharing characteristics with their training data, making OmniObject3D a more objective benchmark.

\noindent\textbf{Comparison on Our Evaluation Datasets.}
Table~\ref{tab:pose} demonstrates the significant performance advantage of FreeSplatter-O over existing baselines on object-centric datasets. MASt3R's reduced effectiveness in this context can be attributed to domain gaps between its training data and background-free rendered images. In scene-level evaluation, FreeSplatter-S matches or exceeds MASt3R's performance, showing superior RRA@$15^\circ$ and TE metrics on ScanNet++ and CO3Dv2. Notably, FreeSplatter-S achieves state-of-the-art performance on the challenging Re10K benchmark despite utilizing a smaller training corpus compared to MASt3R.

\subsection{Ablation Studies}
\label{sec:exp:ablation}

\noindent\textbf{Model Architecture.} 
We analyze architectural choices using a base configuration of 24 transformer layers with patch size 8. Table~\ref{tab:ablation_arch} demonstrates consistent performance improvements on GSO with increased layer count and reduced patch size, attributed to enhanced model capacity and reduced information loss, respectively.

\begin{table}[t]
\scriptsize
\renewcommand{\tabcolsep}{2.7mm}
\renewcommand{\arraystretch}{1.0}
\centering
\resizebox{1.0\columnwidth}{!}{
\begin{tabular}{ccccc}
\toprule
\# Layers ($L$) & Patch Size ($P$) & PSNR $\uparrow$ & SSIM $\uparrow$ & LPIPS $\downarrow$ \\
\midrule
16 & 16 & 25.417 & 0.896 & 0.088 \\
16 & 8 & 28.945 & 0.934 & 0.064 \\  
24 & 16 & 28.622 & 0.927 & 0.063 \\
\textbf{24} & \textbf{8} & \cellcolor{cyan!20}30.443 & \cellcolor{cyan!20}0.945 & \cellcolor{cyan!20}0.055 \\

\bottomrule
\end{tabular}
}
\vspace{-3mm}
\caption{\textbf{Ablation Study on Model Architecture.} The results are evaluated on the GSO dataset with FreeSplatter-O.}
\vspace{-6mm}
\label{tab:ablation_arch}
\end{table}

\noindent\textbf{View Embedding Addition.} 
We evaluate the impact of view embedding addition as formulated in Equation~\ref{eq:embed}. Experiments reveal that assigning $\ve^{\mathrm{ref}}$ to the $j$-th view's tokens and $\ve^{\mathrm{src}}$ to remaining views' tokens enables successful reference view identification and accurate Gaussian reconstruction in the corresponding camera frame. Alternative embedding combinations result in degraded reconstruction quality (details in Section 2.7 of supplementary material).

\noindent\textbf{Number of Input Views.} 
We conduct an experiment on the GSO dataset to illustrate how the number of input views influences the reconstruction quality. Please refer to Figure 13 of the supplementary material for more details. 

\begin{table}[t]
\scriptsize
\renewcommand{\tabcolsep}{2.2mm}
\renewcommand{\arraystretch}{1.0}
\centering
\resizebox{1.0\columnwidth}{!}{
\begin{tabular}{ccccccc}
\toprule
\multirow{2}{*}{$\mathcal{L}_\mathrm{align}$} & 
\multicolumn{3}{c}{\textbf{GSO}} & 
\multicolumn{3}{c}{\textbf{ScanNet++}} \\
\cmidrule(r){2-4} \cmidrule(r){5-7}
& PSNR$\uparrow$ & SSIM$\uparrow$ & LPIPS$\downarrow$ & PSNR$\uparrow$ & SSIM$\uparrow$ & LPIPS$\downarrow$ \\
\midrule
\XSolidBrush & 26.684 & 0.898 & 0.092 & 21.330 & 0.832 & 0.201 \\
\Checkmark & \cellcolor{cyan!20}30.443 & \cellcolor{cyan!20}0.945 & \cellcolor{cyan!20}0.055 & \cellcolor{cyan!20}25.807 & \cellcolor{cyan!20}0.887 & \cellcolor{cyan!20}0.140 \\
\bottomrule
\end{tabular}
}
\vspace{-3mm}
\caption{\textbf{Ablation Study on Pixel-alignment Loss.} The results on GSO and ScanNet++ are evaluated with FreeSplatter-O and FreeSplatter-S, respectively.}
\label{tab:ablation_align}
\vspace{-4mm}
\end{table}

\noindent\textbf{Pixel-Alignment Loss.} 
Ablation on the pixel-alignment loss (Equation~\ref{eq:loss:align}) demonstrates its crucial role in both object and scene-level reconstruction. Its removal leads to significant degradation across all metrics on GSO and ScanNet++ datasets (Table~\ref{tab:ablation_align}). Figure 12 in the supplementary material illustrates how this loss term enhances visual fidelity, with its absence resulting in notable blur artifacts.

\subsection{Applications in 3D AIGC}
\label{sec:exp:application}

FreeSplatter integrates seamlessly into 3D content creation pipelines, offering substantial operational advantages through its pose-free architecture. In contrast, traditional pipelines~\cite{li2024instantd, xu2024instantmesh, tang2024lgm, wang2024crm} require precise alignment between the camera configurations of multi-view diffusion models and the parameters of LRMs, which introduces complexity and potential sources of error. FreeSplatter removes these constraints, enabling direct processing of multi-view images without the need for camera pose information. This streamlined workflow not only reduces generation time for users but also maintains—or even improves—reconstruction quality. In our supplementary material (Section 2.4), we provide comprehensive image-to-3D generation results across a range of multi-view diffusion models, demonstrating that FreeSplatter achieves superior reconstruction performance compared to pose-dependent LRMs and can accurately recover predefined camera parameters from diffusion model outputs.

\section{Conclusion}
FreeSplatter presents a scalable framework for pose-free sparse-view reconstruction. Leveraging a single-stream transformer architecture and unified-frame Gaussian map prediction, the framework delivers both high-fidelity 3D reconstruction and efficient camera pose estimation. Its two specialized variants, designed for object-centric and scene-level reconstruction, achieve superior performance in terms of both reconstruction quality and pose accuracy. Additionally, FreeSplatter shows significant potential in boosting the productivity of downstream applications such as text/image-to-3D content creation, freeing users from the complexities associated with camera pose handling.

{
    \small
    \bibliographystyle{ieeenat_fullname}
    \bibliography{main}
}

\end{document}